\pdfoutput=1
\documentclass[11pt]{article}
\usepackage[margin=1in]{geometry}
\usepackage[T1]{fontenc}
\usepackage{lmodern}
\usepackage{amsmath,amssymb,amsthm,mathtools,bm,booktabs,array,microtype,graphicx}
\usepackage{hyperref}
\hypersetup{colorlinks=true,linkcolor=blue,citecolor=blue,urlcolor=blue}

\newtheorem{theorem}{Theorem}[section]
\newtheorem{lemma}[theorem]{Lemma}
\newtheorem{proposition}[theorem]{Proposition}
\newtheorem{corollary}[theorem]{Corollary}
\theoremstyle{remark}

\newcommand{\E}{\mathbb E}
\newcommand{\Var}{\operatorname{Var}}
\newcommand{\Cov}{\operatorname{Cov}}
\newcommand{\R}{\mathbb R}
\newcommand{\1}{\mathbf 1}

\title{Teacher Supervision, Fixed Representations and Calibration:\\
Exact Risk Calculations for Candidate-Scoring Decision Models}
\author{Agus Sudjianto \and Denis Burakov}
\date{October 2026}

\begin{document}
\maketitle

\begin{abstract}
Candidate-scoring decision models return a probability distribution over a presented set of answers. Public documentation of one such system, Jev, and of an open counterpart, JevAny, describes three training ingredients that determine what these probabilities estimate: soft targets from larger models, a frozen pretrained backbone and temperature calibration. We analyze each ingredient in Gaussian surrogates where the finite-sample risk can be computed exactly. A student trained on teacher targets inherits the teacher's bias, which more student data does not remove and agreement with the teacher does not reveal. A fixed representation lowers the risk only below an explicit sample-size threshold, and it passes on as bias only the part of the teacher's error that lies in the retained directions. In a probit model the exact recalibration map is a temperature scaling. It removes all excess risk only when the teacher's coefficient has the correct direction and otherwise leaves a residual, given in closed form, that calibration diagnostics do not detect. Under Luce's choice axiom, candidate scores are coherent across refinements of an event if and only if they aggregate by log-sum-exp. Simulations agree with the formulas under Gaussian design and show that they understate the risk under heavy-tailed design at small samples. On 231,521 credit card complaints from the Consumer Financial Protection Bureau, where the decision is whether a complaint ends in monetary relief, a student trained on a logistic teacher's probabilities ends at the teacher's decision cost while a student trained on outcomes goes 8\% below it, recalibration of a 30B model read zero-shot corrects its calibration and leaves it worse than the teacher, Turing's calibration check passes on a constant forecast and on a model that ranks at chance, and the 30B model departs from the coherence condition by 17.5 decibans. The probabilities of a candidate-scoring model estimate what its training targets estimate. A pretrained representation can reduce the amount of labeled data required and calibration can correct the scale of the probabilities, but neither corrects a biased teacher, so the probabilities have to be checked against realized outcomes.
\end{abstract}


\section{Introduction}\label{sec:intro}
A candidate-scoring decision model receives a state, a question and a set of candidate answers, and it returns a probability for each candidate. When the candidates are a fixed set of class labels the model is a classifier with a text interface. The probability it returns estimates a conditional distribution, and the training targets determine which distribution that is. The probability that an event occurs, the probability that an annotator selects a candidate and the probability that a reference model assigns to a candidate are different conditional distributions, and a model can estimate one of them well while estimating the others poorly.

Jev is a candidate-scoring model released by TypeSafe \cite{typesafe}. The announcement describes it as a model for structured decisions that returns all outputs in a single query, trained with a method called reinforcement learning for calibrated decisions; for high-cardinality choices it scores candidates independently and then makes an explicit choice. The announcement also states that its workflow evaluations use the predictions of large external models as reference probabilities. The training objective itself is not specified in detail. The JevAny documentation \cite{jevany} describes an openly documented system of the same type. Its released models keep the backbone weights frozen and train low-rank (LoRA, rank 8) adapters together with a scoring head; training uses either a hard target or a normalized soft target distribution over the candidates; and a temperature fitted on a separate calibration set is applied to the candidate scores. Jev also appears as one of the AI judges in the adapted imitation game of Burakov \cite[Chap.~1]{burakov}.

These descriptions identify three mechanisms that bear on the meaning of the output probabilities. Soft targets and reference probabilities make the output of a teacher the quantity that the student learns to reproduce \cite{hinton}. A frozen backbone restricts the student to functions of a fixed representation. Temperature calibration changes the scale of the scores without changing their ranking \cite{guo}. Because the full Jev objective is not public, we do not model Jev. We analyze each mechanism in a Gaussian surrogate in which the finite-sample risk and the calibration curve can be computed exactly, and we state which conclusions depend on the Gaussian assumptions.

Section~\ref{sec:regression} treats least squares with Gaussian random design. The outcome-supervised risk is classical \cite{breiman}; the teacher-target and mixed-supervision risks follow from the same inverse-Wishart calculation and are included for completeness. Section~\ref{sec:rep} gives the exact risk of a student restricted to a fixed linear representation and trained on a mixture of outcomes and teacher targets, in the case where the teacher depends on directions that the representation discards and the two target errors are correlated (Theorem~\ref{th:unified}). From it we obtain the optimal teacher weight and a projection identity for the teacher bias that the compressed student inherits. Section~\ref{sec:probit} turns to a binary outcome under a Gaussian-probit model. It gives the calibration curve of a teacher with an arbitrary coefficient distortion (Theorem~\ref{th:generalcal}), shows that the exact recalibration map is a temperature scaling and gives the residual Brier risk after recalibration in closed form (Corollary~\ref{cor:recal}). Section~\ref{sec:coherence} states the coherence condition for candidate scores, Section~\ref{sec:mc} reports Monte Carlo checks, including designs that violate the Gaussian assumption, and Section~\ref{sec:cfpb} tests the predictions on credit card complaints, where the evidence is text and the true probability is unknown.

\section{Estimands}\label{sec:estimands}
Let $X\in\R^d$ denote the observed predictors, $Y$ a realized outcome and $T$ the output of a teacher. For a binary outcome the target of an outcome-supervised predictor is $\eta(x)=\Pr(Y=1\mid X=x)$, whereas a student trained on teacher outputs estimates $t(x)=\E[T\mid X=x]$. A strictly proper scoring rule is minimized in expectation by the conditional distribution of the training target \cite{gneiting}, so a low proper score against teacher targets establishes that the student reproduces $t$; it does not establish that $t=\eta$.

A candidate-choice model may estimate $\Pr(A=a\mid X=x,Q=q,\mathcal C)$, where $A$ is the candidate selected by an annotator, a reference model or a policy, $Q$ is the question and $\mathcal C$ is the presented candidate set. This probability equals $\Pr(Y=a\mid X=x)$ only under a stated relationship between $A$ and $Y$. A related distinction applies to evidential readings of model scores. Good \cite{good} describes Turing's weight of evidence as the logarithm of a Bayes factor, a ratio of likelihoods under two specified hypotheses. A logit or a log-ratio of softmax outputs has this interpretation only when its numerator and denominator are such likelihoods.

The exact calculations use two surrogates: linear regression with squared error (Sections~\ref{sec:regression} and~\ref{sec:rep}) and a binary probit model (Section~\ref{sec:probit}). The representation is a fixed linear map chosen before the labeled sample is drawn. This corresponds to a frozen backbone with a linear head. Adapting the backbone, for example through LoRA adapters, changes the representation with the labels and lies outside the exact results.

\section{Least squares with Gaussian random design}\label{sec:regression}
\subsection{Outcome supervision}
Let $(X_i,Y_i)$, $i=1,\ldots,n$, be independent with
\begin{equation}\label{eq:dgp}
 X_i\sim N_d(0,\Sigma),\qquad Y_i=X_i^\top\beta+\varepsilon_i,
 \qquad \varepsilon_i\sim N(0,\sigma^2),
\end{equation}
where $\Sigma\succ0$ and the errors are independent of the design and of each other. Let $\mathbf X$ be the $n\times d$ design matrix and $X_0$ an independent test input from the same distribution. For a coefficient estimator $\widehat\theta$ the excess prediction risk is
\begin{equation}\label{eq:risk}
 \mathcal E(\widehat\theta)=\E\big[(X_0^\top\widehat\theta-X_0^\top\beta)^2\big]
 =\E\big[(\widehat\theta-\beta)^\top\Sigma(\widehat\theta-\beta)\big],
\end{equation}
and the mean squared error for a fresh outcome is $\sigma^2+\mathcal E(\widehat\theta)$.

\begin{lemma}[Outcome-supervised least squares {\cite{breiman}}]\label{th:ols}
If $n>d+1$, the estimator $\widehat\beta_O=(\mathbf X^\top\mathbf X)^{-1}\mathbf X^\top\mathbf Y$ satisfies
\begin{equation}\label{eq:olsrisk}
\mathcal E(\widehat\beta_O)=\frac{\sigma^2d}{n-d-1}.
\end{equation}
\end{lemma}
\begin{proof}
Conditionally on $\mathbf X$, $\widehat\beta_O-\beta\sim N(0,\sigma^2(\mathbf X^\top\mathbf X)^{-1})$. Since $\mathbf X^\top\mathbf X\sim W_d(\Sigma,n)$, its inverse has mean $\Sigma^{-1}/(n-d-1)$ for $n>d+1$ \cite{anderson,muirhead}. Taking the expectation of $\sigma^2\operatorname{tr}\{\Sigma(\mathbf X^\top\mathbf X)^{-1}\}$ gives \eqref{eq:olsrisk}.
\end{proof}

\subsection{Teacher supervision}
Suppose a teacher supplies $T_i=X_i^\top(\beta+\delta)+\xi_i$, where $\xi_i\sim N(0,\tau^2)$ is independent of the design. The vector $\delta$ is the systematic teacher error, and the student fits least squares to $T$ in place of $Y$.

\begin{proposition}[Teacher-target risk]\label{th:teacher}
For $n>d+1$ the estimator $\widehat\beta_T=(\mathbf X^\top\mathbf X)^{-1}\mathbf X^\top\mathbf T$ satisfies
\begin{equation}\label{eq:teacherrisk}
\mathcal E(\widehat\beta_T)=\delta^\top\Sigma\delta+\frac{\tau^2d}{n-d-1}.
\end{equation}
Hence $\mathcal E(\widehat\beta_T)<\mathcal E(\widehat\beta_O)$ if and only if
\begin{equation}\label{eq:criterion}
\delta^\top\Sigma\delta<\frac{d(\sigma^2-\tau^2)}{n-d-1}.
\end{equation}
\end{proposition}
\begin{proof}
The coefficient error is $\widehat\beta_T-\beta=\delta+(\mathbf X^\top\mathbf X)^{-1}\mathbf X^\top\bm\xi$. The second term has conditional mean zero, so the cross term vanishes, and its $\Sigma$-weighted second moment follows from the calculation in Lemma~\ref{th:ols}.
\end{proof}

A fixed $\delta\neq0$ leaves the error floor $\delta^\top\Sigma\delta$ as $n\to\infty$, whereas $\mathcal E(\widehat\beta_O)\to0$ for fixed $d$. A teacher with $\tau^2<\sigma^2$ therefore gives lower risk than outcome supervision only for $n$ below the crossover $d+1+d(\sigma^2-\tau^2)/\delta^\top\Sigma\delta$.

\subsection{Mixed supervision}
Let $\widehat\beta_\lambda$ minimize $(1-\lambda)\|\mathbf Y-\mathbf Xb\|^2+\lambda\|\mathbf T-\mathbf Xb\|^2$ for $\lambda\in[0,1]$. Allow the pairs $(\varepsilon_i,\xi_i)$ to be jointly Gaussian with variances $\sigma^2,\tau^2$ and covariance $\omega$, independent across $i$ and of the design. Put $B=\delta^\top\Sigma\delta$ and $C=d/(n-d-1)$.

\begin{proposition}[Mixed supervision with correlated errors]\label{th:mix}
For $n>d+1$,
\begin{equation}\label{eq:mixrisk}
\mathcal E(\widehat\beta_\lambda)=\lambda^2B+C\big[(1-\lambda)^2\sigma^2+\lambda^2\tau^2+2\lambda(1-\lambda)\omega\big].
\end{equation}
If $B+C(\sigma^2+\tau^2-2\omega)>0$, the unconstrained minimizer is
\begin{equation}\label{eq:lambda}
\lambda_{\mathrm{unc}}=\frac{C(\sigma^2-\omega)}{B+C(\sigma^2+\tau^2-2\omega)},
\end{equation}
and the minimizer over $[0,1]$ is $\lambda^*=\min\{1,\max\{0,\lambda_{\mathrm{unc}}\}\}$. In particular $\lambda_{\mathrm{unc}}<0$ if and only if $\omega>\sigma^2$, and $\lambda_{\mathrm{unc}}>1$ if and only if $\omega>\tau^2+B/C$. For $\omega=0$ and $\sigma^2>0$, $\lambda^*=C\sigma^2/\{B+C(\sigma^2+\tau^2)\}\in(0,1]$.
\end{proposition}
\begin{proof}
The objective is least squares on the target $(1-\lambda)\mathbf Y+\lambda\mathbf T$, so the coefficient error is $\lambda\delta+(\mathbf X^\top\mathbf X)^{-1}\mathbf X^\top\{(1-\lambda)\bm\varepsilon+\lambda\bm\xi\}$. The noise term has conditional mean zero and conditional covariance $\{(1-\lambda)^2\sigma^2+\lambda^2\tau^2+2\lambda(1-\lambda)\omega\}(\mathbf X^\top\mathbf X)^{-1}$, which gives \eqref{eq:mixrisk}. The risk is a quadratic in $\lambda$ with leading coefficient $B+C\Var(\varepsilon-\xi)\ge0$; setting the derivative to zero gives \eqref{eq:lambda}, and the projection onto $[0,1]$ minimizes a convex quadratic over the interval. The two boundary conditions follow by comparing the numerator of \eqref{eq:lambda} with zero and with the denominator.
\end{proof}

Since $\omega\le\sigma\tau\le\max\{\sigma^2,\tau^2\}$, at most one of the two boundary conditions can hold. If the leading coefficient vanishes, then $B=0$ and $\varepsilon=\xi$ almost surely, the linear coefficient $C(\sigma^2-\omega)$ also vanishes and the risk does not depend on $\lambda$. For $B>0$ the optimal weight tends to zero as $n\to\infty$ because $C\to0$.

\section{Fixed representations}\label{sec:rep}
\subsection{Outcome supervision through a representation}
Let $A\in\R^{r\times d}$ have full row rank, $1\le r\le d$, and let $Z=AX$ with covariance $C_A=A\Sigma A^\top$. The map $A$ is fixed before the labeled sample is drawn. The best linear predictor of $Y$ from $Z$ has coefficient $w_*=C_A^{-1}A\Sigma\beta$, and the part of the signal that $Z$ does not capture has coefficient and variance
\begin{equation}\label{eq:resid}
 u_\beta=\beta-A^\top w_*,\qquad B_A=u_\beta^\top\Sigma u_\beta
 =\beta^\top\big(\Sigma-\Sigma A^\top C_A^{-1}A\Sigma\big)\beta .
\end{equation}
We call $B_A$ the representation loss.

\begin{proposition}[Risk through a fixed representation]\label{th:rep}
Let $\widehat w_O$ be the least-squares coefficient of $Y$ on $Z$ with $n>r+1$. Its excess risk relative to $X^\top\beta$ is
\begin{equation}\label{eq:reprisk}
\mathcal E_A=B_A+\frac{r(\sigma^2+B_A)}{n-r-1}.
\end{equation}
\end{proposition}
\begin{proof}
Since $\Cov(Z,X^\top u_\beta)=A\Sigma u_\beta=0$ and $(Z,X^\top u_\beta)$ is jointly Gaussian, $X^\top u_\beta$ is independent of $Z$ with variance $B_A$. Therefore $Y=Z^\top w_*+e$ with $e=X^\top u_\beta+\varepsilon$ independent of $Z$ and $\Var(e)=\sigma^2+B_A$. The prediction error at a fresh point is $Z_0^\top(\widehat w_O-w_*)-X_0^\top u_\beta$; the two terms are uncorrelated, the second has second moment $B_A$ and the first has expected square $r(\sigma^2+B_A)/(n-r-1)$ by Lemma~\ref{th:ols} applied in dimension $r$.
\end{proof}

\begin{corollary}[Compression threshold]\label{cor:crossover}
For $n>d+1$, $\mathcal E_A<\mathcal E(\widehat\beta_O)$ if and only if
\begin{equation}\label{eq:crossover}
 B_A<\frac{\sigma^2(d-r)}{n-d-1}.
\end{equation}
\end{corollary}
\begin{proof}
Write $\mathcal E_A=\{(n-1)B_A+r\sigma^2\}/(n-r-1)$ and compare it with $\sigma^2d/(n-d-1)$. The identity $d(n-r-1)-r(n-d-1)=(d-r)(n-1)$ reduces the comparison to \eqref{eq:crossover}.
\end{proof}

The threshold decreases to zero as $n$ grows, and it is zero for $r=d$. A representation with $B_A>0$ is therefore preferable to the full predictors only below the sample size $d+1+\sigma^2(d-r)/B_A$.

\subsection{Mixed supervision through a representation}
A teacher may depend on directions of $X$ that the student's representation discards. Let $\gamma=\beta+\delta$ and $T=X^\top\gamma+\xi$, and define
\begin{equation}
 w_\gamma=C_A^{-1}A\Sigma\gamma,\qquad
 u_\gamma=\gamma-A^\top w_\gamma,
 \qquad D_A=(w_\gamma-w_*)^\top C_A(w_\gamma-w_*),
\end{equation}
together with $G_A=u_\gamma^\top\Sigma u_\gamma$ and $S_A=u_\beta^\top\Sigma u_\gamma$. The error pairs $(\varepsilon_i,\xi_i)$ are as in Proposition~\ref{th:mix}. The student fits a linear head on $Z$ to the targets $V_\lambda=(1-\lambda)Y+\lambda T$.

\begin{theorem}[Exact risk with representation and mixed supervision]\label{th:unified}
For $n>r+1$ the least-squares head $\widehat w_\lambda$ of $V_\lambda$ on $Z$ satisfies
\begin{equation}\label{eq:unified}
\begin{aligned}
\mathcal E_{A,\lambda}={}& B_A+\lambda^2D_A+\frac{r}{n-r-1}\Big[ (1-\lambda)^2(B_A+\sigma^2) \\
 &+\lambda^2(G_A+\tau^2)+2\lambda(1-\lambda)(S_A+\omega)\Big].
\end{aligned}
\end{equation}
\end{theorem}
\begin{proof}
Let $w_\lambda=(1-\lambda)w_*+\lambda w_\gamma$. Then $V_\lambda=Z^\top w_\lambda+e_\lambda$ with
$e_\lambda=(1-\lambda)(X^\top u_\beta+\varepsilon)+\lambda(X^\top u_\gamma+\xi)$.
Both $X^\top u_\beta$ and $X^\top u_\gamma$ are uncorrelated with $Z$, because $A\Sigma u_\beta=A\Sigma u_\gamma=0$, and the errors are independent of $X$. By joint Gaussianity $e_\lambda$ is independent of $Z$, and its variance is the bracket in \eqref{eq:unified}, since $\Cov(X^\top u_\beta+\varepsilon,X^\top u_\gamma+\xi)=S_A+\omega$. The prediction error at a fresh point is
\[
Z_0^\top(\widehat w_\lambda-w_\lambda)+\lambda Z_0^\top(w_\gamma-w_*)-X_0^\top u_\beta .
\]
The last term is uncorrelated with $Z_0$ and independent of the training sample, and the first term has mean zero given the training design. The three terms are therefore uncorrelated, with expected squares equal to the inverse-Wishart variance term, $\lambda^2D_A$ and $B_A$.
\end{proof}

Equation~\eqref{eq:unified} is not the sum of a representation loss, a teacher error and an estimation variance that could each be computed separately. The term $S_A$ couples the outcome and teacher residuals, and the unrepresented teacher signal $G_A$ enters the variance term rather than the bias term.

\begin{corollary}[Reduction to Proposition~\ref{th:mix}]
If $u_\beta=u_\gamma=0$, then $B_A=G_A=S_A=0$ and $D_A=\delta^\top\Sigma\delta$, so \eqref{eq:unified} coincides with \eqref{eq:mixrisk} with $d$ replaced by $r$.
\end{corollary}

\begin{corollary}[Projected teacher bias]\label{cor:project}
With $\delta=\gamma-\beta$,
\begin{equation}\label{eq:projectbias}
 D_A=\delta^\top\Sigma A^\top (A\Sigma A^\top)^{-1}A\Sigma\delta
 \leq\delta^\top\Sigma\delta .
\end{equation}
Equality holds if and only if $\Sigma^{1/2}\delta$ lies in the row space of $A\Sigma^{1/2}$, and $D_A=0$ if and only if $A\Sigma\delta=0$.
\end{corollary}
\begin{proof}
Since $w_\gamma-w_*=C_A^{-1}A\Sigma\delta$, the first equality follows from the definition of $D_A$. The matrix $P=\Sigma^{1/2}A^\top(A\Sigma A^\top)^{-1}A\Sigma^{1/2}$ is the orthogonal projector onto the row space of $A\Sigma^{1/2}$, and $D_A=\|P\Sigma^{1/2}\delta\|^2\leq\|\Sigma^{1/2}\delta\|^2$, with equality exactly when $\Sigma^{1/2}\delta$ lies in the range of $P$. Finally $P\Sigma^{1/2}\delta=0$ if and only if $(A\Sigma A^\top)^{-1}A\Sigma\delta=0$, that is, $A\Sigma\delta=0$.
\end{proof}

The projection removes the component of the teacher bias that the student cannot represent from the coefficient bias. That component remains in the training targets, however, and it contributes to $G_A$ and $S_A$ and hence to the variance of the fitted head. Compression thus converts part of the systematic teacher error into target noise.

\begin{corollary}[Optimal teacher weight under compression]\label{cor:unifiedopt}
Let $C_r=r/(n-r-1)$, $U=B_A+\sigma^2$, $V=G_A+\tau^2$ and $H=S_A+\omega$. If $D_A+C_r(U+V-2H)>0$, the unconstrained minimizer of \eqref{eq:unified} is
\begin{equation}\label{eq:lambdaunified}
 \lambda_{A,\mathrm{unc}}=\frac{C_r(U-H)}{D_A+C_r(U+V-2H)},
\end{equation}
and the minimizer over $[0,1]$ is its projection onto that interval. If the denominator is zero, the risk does not depend on $\lambda$.
\end{corollary}
\begin{proof}
Expanding \eqref{eq:unified} gives $B_A+C_rU-2C_r(U-H)\lambda+\{D_A+C_r(U+V-2H)\}\lambda^2$. The leading coefficient is nonnegative because $U+V-2H=\Var\{(X^\top u_\beta+\varepsilon)-(X^\top u_\gamma+\xi)\}$. If it vanishes, the two residuals are almost surely equal, so $U=V=H$ and the linear coefficient vanishes as well.
\end{proof}

Compression changes $B_A$, $D_A$, $G_A$, $S_A$ and $C_r$ together, so it has no fixed effect on the preferred teacher weight. In the example of Section~\ref{sec:mc} the optimal weights with and without compression are $0.4218$ and $0.4221$.

\section{Binary outcomes under a Gaussian-probit model}\label{sec:probit}
\subsection{Excess Brier risk and calibration of a distorted teacher}
Let $X\sim N_d(0,\Sigma)$, $\varepsilon\sim N(0,\sigma^2)$ independent of $X$ and $Y=\1\{X^\top\beta+\varepsilon>0\}$. Then
\begin{equation}\label{eq:probit}
\eta(x)=\Pr(Y=1\mid X=x)=\Phi(a^\top x),\qquad a=\beta/\sigma .
\end{equation}
A teacher reports $t(x)=\Phi(b^\top x)$. In the population, minimizing cross-entropy against these soft targets returns $t$ whenever the student's function class contains it, and the excess Brier risk of $t$ relative to $\eta$ is $\E[\{t(X)-\eta(X)\}^2]$. Throughout we write $v_a=a^\top\Sigma a$, $v_b=b^\top\Sigma b$ and $c=a^\top\Sigma b$.

\begin{proposition}[Excess Brier risk of a probit teacher]\label{th:probit}
\begin{equation}\label{eq:arcsin}
\E[\{t(X)-\eta(X)\}^2]=\frac{1}{2\pi}\left[
\arcsin\frac{v_a}{1+v_a}+\arcsin\frac{v_b}{1+v_b}
-2\arcsin\frac{c}{\sqrt{(1+v_a)(1+v_b)}}\right].
\end{equation}
\end{proposition}
\begin{proof}
Let $(U,V)$ be centered jointly Gaussian with variances $s_U^2,s_V^2$ and covariance $s_{UV}$, and let $E_1,E_2$ be independent standard normal variables independent of $(U,V)$. Then $\E[\Phi(U)\Phi(V)]=\Pr(U-E_1\ge0,\,V-E_2\ge0)$, and the bivariate orthant probability \cite{sheppard} gives
$\E[\Phi(U)\Phi(V)]=\tfrac14+(2\pi)^{-1}\arcsin\{s_{UV}/\sqrt{(1+s_U^2)(1+s_V^2)}\}$.
Applying this to the pairs $(a^\top X,a^\top X)$, $(b^\top X,b^\top X)$ and $(a^\top X,b^\top X)$ and expanding the square gives \eqref{eq:arcsin}; the constants $\tfrac14$ cancel.
\end{proof}

\begin{proposition}[Calibration under coefficient scaling]\label{th:scale}
Suppose $v_a>0$ and $b=ka$ with $k>0$, and let $P=\Phi(ka^\top X)$. For $p\in(0,1)$,
\begin{equation}\label{eq:calibration}
\Pr(Y=1\mid P=p)=\Phi\left(\frac{\Phi^{-1}(p)}{k}\right).
\end{equation}
The forecast $P$ ranks observations as $\eta(X)$ does and has the same AUC, and it is calibrated if and only if $k=1$. Dividing the probit score by the temperature $k$ recovers $\eta(X)$ exactly.
\end{proposition}
\begin{proof}
The map $u\mapsto\Phi(ku)$ is strictly increasing, so $P=p$ is equivalent to $a^\top X=\Phi^{-1}(p)/k$. Since $P$ is a function of $X$, $\Pr(Y=1\mid P)=\E[\eta(X)\mid P]=\Phi(a^\top X)$, which gives \eqref{eq:calibration}. A strictly increasing transformation leaves the ranking unchanged, and \eqref{eq:calibration} equals $p$ for all $p$ only when $k=1$.
\end{proof}

\begin{theorem}[Calibration under arbitrary coefficient distortion]\label{th:generalcal}
Assume $v_b>0$ and let $P=\Phi(b^\top X)$. Then
\begin{equation}\label{eq:generalcal}
 \Pr(Y=1\mid P=p)=\Phi\big(\kappa\,\Phi^{-1}(p)\big),\qquad
 \kappa=\frac{c/v_b}{\sqrt{1+v_a-c^2/v_b}} .
\end{equation}
The result covers rotations of the coefficient vector as well as changes of its length, and it reduces to \eqref{eq:calibration} when $b=ka$, $k>0$.
\end{theorem}
\begin{proof}
Conditionally on $b^\top X=s$, the variable $a^\top X$ is Gaussian with mean $cs/v_b$ and variance $v_a-c^2/v_b$. For $U\sim N(m,v)$ and an independent standard normal $E$, $\E[\Phi(U)]=\Pr(E\le U)=\Phi(m/\sqrt{1+v})$. Substituting $s=\Phi^{-1}(p)$ gives \eqref{eq:generalcal}. For $b=ka$ one has $c/v_b=1/k$ and $c^2/v_b=v_a$, so $\kappa=1/k$.
\end{proof}

The calibration curve \eqref{eq:generalcal} is linear on the probit scale with zero intercept. When $c=0$, the scores $a^\top X$ and $b^\top X$ are independent, $P$ carries no information about $Y$ and the curve is constant at $\tfrac12$.

\subsection{Recalibration}
\begin{corollary}[Residual risk after exact recalibration]\label{cor:recal}
Under the assumptions of Theorem~\ref{th:generalcal}, the Brier-optimal forecast among functions of $P$ is
\begin{equation}
P_{\mathrm{cal}}=\E[Y\mid P]=\Phi(\kappa\, b^\top X),
\end{equation}
that is, the teacher's probit score divided by the temperature $1/\kappa$ when $c>0$. Its excess Brier risk is
\begin{equation}\label{eq:residual}
\E[\{P_{\mathrm{cal}}-\eta(X)\}^2]=\frac{1}{2\pi}\left[\arcsin\frac{v_a}{1+v_a}-\arcsin\frac{c^2}{v_b(1+v_a)}\right],
\end{equation}
which is zero if and only if $b$ is a nonzero multiple of $a$. For $c>0$ the recalibration map is strictly increasing and preserves the ranking; for $c<0$ it is strictly decreasing and reverses the ranking; for $c=0$ the recalibrated forecast is constant at $\tfrac12$.
\end{corollary}
\begin{proof}
The first display follows from Theorem~\ref{th:generalcal}, and the conditional expectation minimizes $\E[\{f(P)-Y\}^2]$ over functions $f$. Because $P_{\mathrm{cal}}=\E[\eta(X)\mid P]$, the cross term in the expansion of the square equals $\E[P_{\mathrm{cal}}^2]$, so $\E[(P_{\mathrm{cal}}-\eta)^2]=\E[\eta^2]-\E[P_{\mathrm{cal}}^2]$. The orthant identity in the proof of Proposition~\ref{th:probit} gives $\E[\eta^2]=\tfrac14+(2\pi)^{-1}\arcsin\{v_a/(1+v_a)\}$ and $\E[P_{\mathrm{cal}}^2]=\tfrac14+(2\pi)^{-1}\arcsin\{\kappa^2v_b/(1+\kappa^2v_b)\}$. From the definition of $\kappa$, $1+\kappa^2v_b=(1+v_a)/(1+v_a-c^2/v_b)$, so $\kappa^2v_b/(1+\kappa^2v_b)=c^2/\{v_b(1+v_a)\}$, which gives \eqref{eq:residual}. The difference is zero if and only if $c^2=v_av_b$, which by the Cauchy--Schwarz inequality in the $\Sigma$ inner product holds if and only if $b$ and $a$ are collinear. The statements about monotonicity follow from the sign of $\kappa$.
\end{proof}

Temperature scaling therefore corrects a teacher whose coefficient has the right direction and the wrong length, but it leaves the excess \eqref{eq:residual} when the direction is wrong. Any function of $P$ has resolution, in the sense of Murphy's decomposition of the Brier score \cite{murphy}, no greater than that of $\E[Y\mid P]$, so no recalibration of $P$ can recover the information about $a^\top X$ that is lost in projecting onto $b^\top X$. The statement that the exact map is a temperature scaling depends on the probit link. Under a logistic link the conditional mean of a logistic function of a Gaussian variable is not itself a logistic function of that variable, and temperature scaling on the logit scale \cite{guo} is then an approximation to the recalibration map.

\subsection{Evaluation against teacher probabilities}
When a student $f$ is evaluated against reference probabilities $t$ instead of outcomes, the quantity measured is $\E[\{f(X)-t(X)\}^2]$. The triangle inequality in $L^2$ gives
\begin{equation}\label{eq:triangle}
\Big|\,\E[\{f(X)-\eta(X)\}^2]^{1/2}-\E[\{t(X)-\eta(X)\}^2]^{1/2}\Big|\le\E[\{f(X)-t(X)\}^2]^{1/2}.
\end{equation}
Agreement with the teacher therefore bounds the student's excess Brier risk only to within the teacher's own excess, which is given by \eqref{eq:arcsin} in the probit model. A student that reproduces the teacher exactly inherits that excess.

\subsection{Information retained by a representation}
For $Z=AX$, Gaussian conditioning gives $X^\top\beta\mid Z=z\sim N(z^\top w_*,B_A)$ with $w_*$ and $B_A$ as in \eqref{eq:resid}. Therefore
\begin{equation}\label{eq:probitrep}
\Pr(Y=1\mid Z=z)=\Phi\left(\frac{z^\top w_*}{\sqrt{\sigma^2+B_A}}\right),
\end{equation}
so the discarded linear signal acts as additional latent noise. Equation~\eqref{eq:probitrep} describes the best forecast available from the representation. It is not a finite-sample risk for a fitted probit model.

\section{Coherence of candidate scores}\label{sec:coherence}
In fixed-label classification the event, and hence the target probability, does not depend on how the candidates are presented. For a candidate-scoring model the wording and membership of the candidate set can change the output. A change is legitimate when it reflects a change in the event being asked about, and it is an inconsistency when the same event receives incompatible probabilities.

Let $q_{\mathcal P}(F\mid x)$ denote the probability that the model assigns to candidate $F$ when the mutually exclusive and exhaustive candidate set $\mathcal P$ is presented with input $x$. Let $E=E_1\cup E_2$ with $E_1\cap E_2=\emptyset$, and let $E^c$ be the complement of $E$. The model is coherent under this refinement if
\begin{equation}\label{eq:coherence}
q_{\{E,E^c\}}(E\mid x)=q_{\{E_1,E_2,E^c\}}(E_1\mid x)+q_{\{E_1,E_2,E^c\}}(E_2\mid x).
\end{equation}
Similarly, a relabeling or reordering of the candidates should permute the probability vector without changing the probability of any event. A softmax layer guarantees normalization within each presented set, but it does not impose \eqref{eq:coherence} across different sets.

\begin{proposition}[Coherence of independently computed scores]\label{prop:coherence}
Suppose each candidate event $F$ receives a positive weight $w(F)=\exp s(F)$ that does not depend on the other candidates presented, and the probabilities are obtained by normalizing the weights over the presented set. Then \eqref{eq:coherence} holds if and only if
\begin{equation}\label{eq:logsum}
\exp s(E)=\exp s(E_1)+\exp s(E_2),\quad\text{equivalently}\quad s(E)=\log\{\exp s(E_1)+\exp s(E_2)\}.
\end{equation}
\end{proposition}
\begin{proof}
Write $u=w(E)$, $v=w(E_1)+w(E_2)$ and $h=w(E^c)>0$. Condition \eqref{eq:coherence} reads $u/(u+h)=v/(v+h)$, which is equivalent to $uh=vh$ and hence to $u=v$.
\end{proof}

The normalization in Proposition~\ref{prop:coherence} is Luce's choice model \cite{luce}, and the assumption that weights do not depend on the other candidates is his choice axiom. Under that assumption, invariance to reordering holds automatically, and \eqref{eq:logsum} is the additional condition required for refinement. The Jev announcement describes independent scoring of candidates for high-cardinality choices \cite{typesafe}. If candidate scores are instead computed within a shared context that contains all candidates, they can depend on the other candidates and on their order, and both reordering invariance and \eqref{eq:coherence} become empirical questions that can be tested by posing the same event under different candidate sets.

\section{Monte Carlo experiments}\label{sec:mc}
\subsection{Design}
All experiments use the parameters in Table~\ref{tab:params}. For the regression results, $\Sigma=I_5$ and $A=[I_3\;0]$, so the test-point expectation in \eqref{eq:risk} is available in closed form given the fitted coefficients, and the Monte Carlo average is taken over $20{,}000$ independent training samples per configuration. These parameters give $B=0.2378$, $B_A=0.0800$, $D_A=0.0753$, $G_A=0.4625$ and $S_A=0.1900$. For the probit results, $\Sigma=I_2$, $a=(1,0)^\top$ and $b=(\cos\theta,\sin\theta)^\top$, so that $v_a=v_b=1$ and $c=\cos\theta$. Calibration curves are estimated from $2\times10^6$ simulated binary outcomes per angle in 20 equal-width forecast bins, and Brier excesses are estimated from $10^8$ draws per angle. The code is in \texttt{experiments.py}, which runs on a single GPU.

\begin{table}[ht]
\centering\small
\caption{Simulation parameters.}\label{tab:params}\vspace{6pt}
\begin{tabular}{p{0.27\linewidth}p{0.65\linewidth}}
\toprule
Quantity & Value \\
\midrule
Design & $d=5$, $r=3$, $\Sigma=I_5$, $A=[I_3\;0]$ \\
Outcome coefficient & $\beta=(0.9,-0.6,0.5,0.2,-0.2)^\top$ \\
Teacher distortion & $\delta=(0.22,-0.13,0.10,0.35,-0.20)^\top$ \\
Errors & $\sigma^2=1.3$, $\tau^2=0.7$, $\omega=0.22$ \\
Sample-size sweep & $n\in\{12,15,20,27,40,60,95,150,250\}$; $\lambda=0.45$ for the mixed head \\
Teacher-weight sweep & $n=27$; 31 equally spaced values of $\lambda$ in $[0,1]$ \\
Replications & $20{,}000$ training samples per configuration \\
Probit calibration & $\theta\in\{0^\circ,30^\circ,60^\circ\}$ \\
Random seed & $20261016$ (PyTorch generator) \\
\bottomrule
\end{tabular}
\end{table}

\subsection{Gaussian design}
\begin{figure}[ht]
\centering
\includegraphics[width=\linewidth]{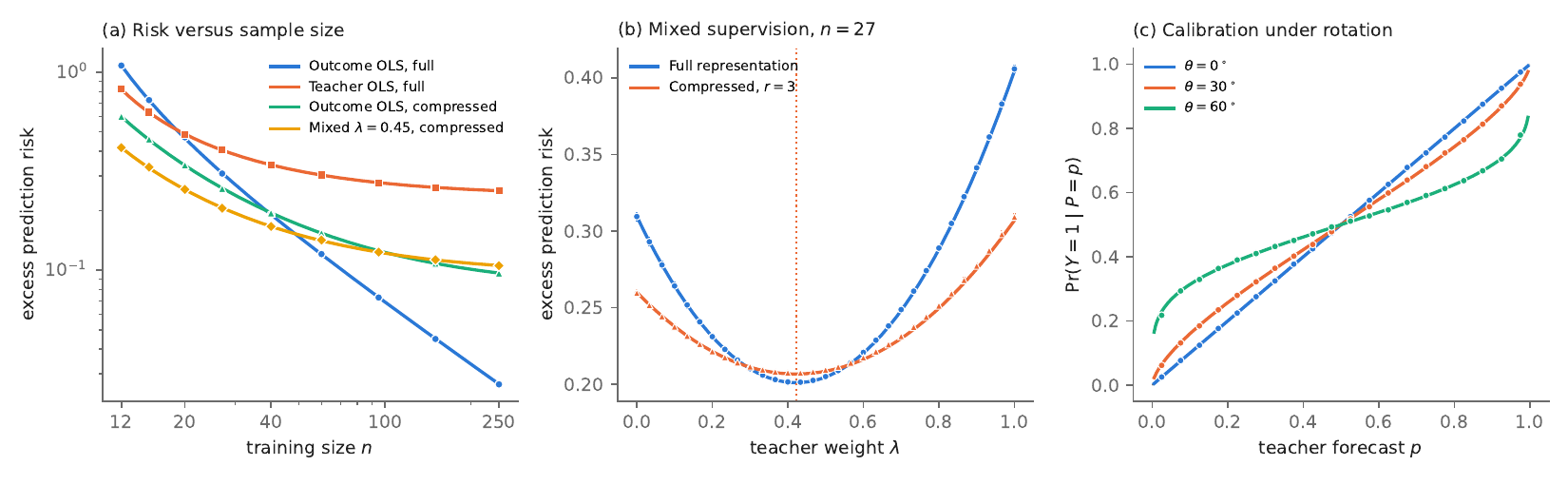}
\caption{Exact formulas (curves) and Monte Carlo estimates (points with 95\% intervals, mostly smaller than the markers). (a) Excess prediction risk against training size for outcome and teacher supervision with all predictors, outcome supervision through the compressed representation and mixed supervision with $\lambda=0.45$ through the compressed representation. (b) Excess risk against teacher weight at $n=27$; dotted lines mark the minimizers from Proposition~\ref{th:mix} and Corollary~\ref{cor:unifiedopt}. (c) Calibration curves of rotated probit teachers (Theorem~\ref{th:generalcal}) and binned event frequencies; the dashed line is the diagonal.}\label{fig:mc}
\end{figure}

Figure~\ref{fig:mc} compares the formulas with simulation. In panel~(a), 34 of the 36 Monte Carlo means lie within their 95\% intervals of the formulas \eqref{eq:olsrisk}, \eqref{eq:teacherrisk}, \eqref{eq:reprisk} and \eqref{eq:unified}; the two exceptions lie at $1.01$ and $1.49$ interval half-widths. Teacher supervision has lower risk than outcome supervision below $n=18.6$, and the compressed outcome model has lower risk than the full model below $n=38.5$, as predicted by \eqref{eq:criterion} and \eqref{eq:crossover}. The teacher-trained model approaches its floor $B=0.2378$, and the compressed models approach $B_A=0.08$ and $B_A+\lambda^2D_A=0.0952$. In panel~(b) all 62 Monte Carlo means lie within their 95\% intervals. The formulas give optimal weights $\lambda^*=0.4221$ for the full representation and $0.4218$ for the compressed one, and the grid minimizers are both $0.433$, the grid point nearest to them. At $n=27$ the minimum risk is $0.2010$ with the full representation and $0.2066$ with the compressed one, so compression does not reduce the risk of the optimally mixed estimator at this sample size. In panel~(c) the binned event frequencies follow the calibration curves of Theorem~\ref{th:generalcal}, with $\kappa=1$, $0.775$ and $0.378$.

Table~\ref{tab:brier} reports the Brier excesses. The simulated values agree with Proposition~\ref{th:probit} and Corollary~\ref{cor:recal} to within $1.3$ standard errors, and two-dimensional Gauss--Hermite quadrature reproduces both formulas to $10^{-15}$. At $\theta=60^\circ$ recalibration reduces the excess Brier risk from $0.0862$ to $0.0634$, and the remainder is the cost of the wrong coefficient direction.

\begin{table}[ht]
\centering\small
\caption{Excess Brier risk relative to $\eta$ for rotated probit teachers, before and after exact recalibration. Monte Carlo standard errors in parentheses ($10^8$ draws per angle).}\label{tab:brier}\vspace{6pt}
\begin{tabular}{rcccc}
\toprule
& \multicolumn{2}{c}{Teacher forecast $P$} & \multicolumn{2}{c}{Recalibrated $P_{\mathrm{cal}}$} \\
\cmidrule(lr){2-3}\cmidrule(lr){4-5}
$\theta$ & Formula \eqref{eq:arcsin} & Monte Carlo & Formula \eqref{eq:residual} & Monte Carlo \\
\midrule
$0^\circ$ & $0$ & $0$ & $0$ & $0$ \\
$30^\circ$ & $0.024117$ & $0.024121\;(0.000004)$ & $0.022155$ & $0.022158\;(0.000003)$ \\
$60^\circ$ & $0.086236$ & $0.086223\;(0.000011)$ & $0.063387$ & $0.063378\;(0.000007)$ \\
\bottomrule
\end{tabular}
\end{table}

\subsection{Non-Gaussian design}
The inverse-Wishart moments and the independence of the residual from $Z$ both use the Gaussian design. To measure the effect of departing from it, we drew $X=\Sigma^{1/2}W$ with an equicorrelated $\Sigma$ ($\Sigma_{jj}=1$, $\Sigma_{jk}=0.5$) and three laws for $W$, each with identity covariance: standard Gaussian, independent uniform coordinates and a multivariate $t$ distribution with five degrees of freedom. The covariance of $X$ is the same under all three laws, so the formulas give the same values. Under the uniform and multivariate-$t$ laws the residual $X^\top u_\beta$ remains uncorrelated with $Z$ but is no longer independent of it. Table~\ref{tab:robust} reports the ratio of the simulated risk to the formula for outcome supervision with all predictors, \eqref{eq:olsrisk}, and for the compressed mixed head with $\lambda=0.45$, \eqref{eq:unified}.

\begin{table}[ht]
\centering\small
\caption{Ratio of simulated excess risk to the Gaussian-design formula, equicorrelated $\Sigma$ with $\rho=0.5$, $20{,}000$ training samples per entry. The 95\% Monte Carlo interval half-widths of the ratios are at most $0.022$.}\label{tab:robust}\vspace{6pt}
\begin{tabular}{lcccccc}
\toprule
& \multicolumn{3}{c}{Outcome OLS, all predictors} & \multicolumn{3}{c}{Mixed head, compressed} \\
\cmidrule(lr){2-4}\cmidrule(lr){5-7}
Design law & $n=12$ & $n=27$ & $n=95$ & $n=12$ & $n=27$ & $n=95$ \\
\midrule
Gaussian & $1.005$ & $1.003$ & $1.000$ & $1.011$ & $1.006$ & $1.000$ \\
Uniform coordinates & $0.903$ & $0.951$ & $0.979$ & $0.928$ & $0.971$ & $0.994$ \\
Multivariate $t_5$ & $1.354$ & $1.197$ & $1.070$ & $1.233$ & $1.127$ & $1.054$ \\
\bottomrule
\end{tabular}
\end{table}

The uniform design, whose coordinates have lighter tails than the Gaussian, gives risks below the formulas, and the multivariate-$t$ design, a scale mixture of Gaussians with heavier tails, gives risks above them. The discrepancies shrink with $n$. They are larger for the full model, whose risk consists entirely of estimation variance, than for the compressed head, whose risk includes the bias terms $B_A$ and $\lambda^2D_A$ that do not depend on the design law. At $n=95$ the formulas are within $7\%$ of simulation under both non-Gaussian laws, and at $n=12$ they understate the risk by $26\%$ of its simulated value under the multivariate-$t$ law.

\section{Application: relief decisions on credit card complaints}\label{sec:cfpb}
We tested the predictions of Sections~\ref{sec:regression}--\ref{sec:coherence} on a credit decision in which the evidence is text and the true probability is unknown, and the direction of every prediction held. When a consumer complains about a credit card, the issuer closes the complaint with monetary relief, with non-monetary relief or with an explanation. A bank can estimate the probability of monetary relief from the complaint and escalate the likely cases. Missing such a case costs more than reviewing one that did not need it, so the decision is taken at a low cutoff.

\subsection{Data and models}
The Consumer Complaint Database of the Consumer Financial Protection Bureau (CFPB) publishes complaints with the consumer's narrative where the consumer consented, with personal details removed \cite{cfpb}. The CFPB announced on 14 August 2026 that it would stop publishing narratives; we use the archive of all narratives received up to that date \cite{cfpbarchive}. We keep credit card complaints closed with one of the three responses and split them by the year received: 37,520 from 2015--2018 train the teacher, 97,835 from 2019--2023 train the students, 34,957 from 2024 serve for calibration and 61,209 from 2025--2026 for testing, after removing 1,088 narratives that occur in more than one split. These are pools. Of the 97,835 student complaints, 4,000 are held out for early stopping and each student is trained on $n=64{,}000$ drawn from the other 93,835. The recalibration curves are fitted on 5,000 of the 34,957 calibration complaints, and every model is evaluated on the same 10,000 of the 61,209 test complaints, a fixed set that makes the bootstrap comparisons paired and keeps the cost of scoring the students down. The share with monetary relief falls from 0.206 in the teacher years to 0.166 in the test years. Every model sees the company, the issue and the narrative. The label is the company's decision, and the models predict that decision, as a triage model would.

\begin{figure}[ht]
\centering
\includegraphics[width=\linewidth]{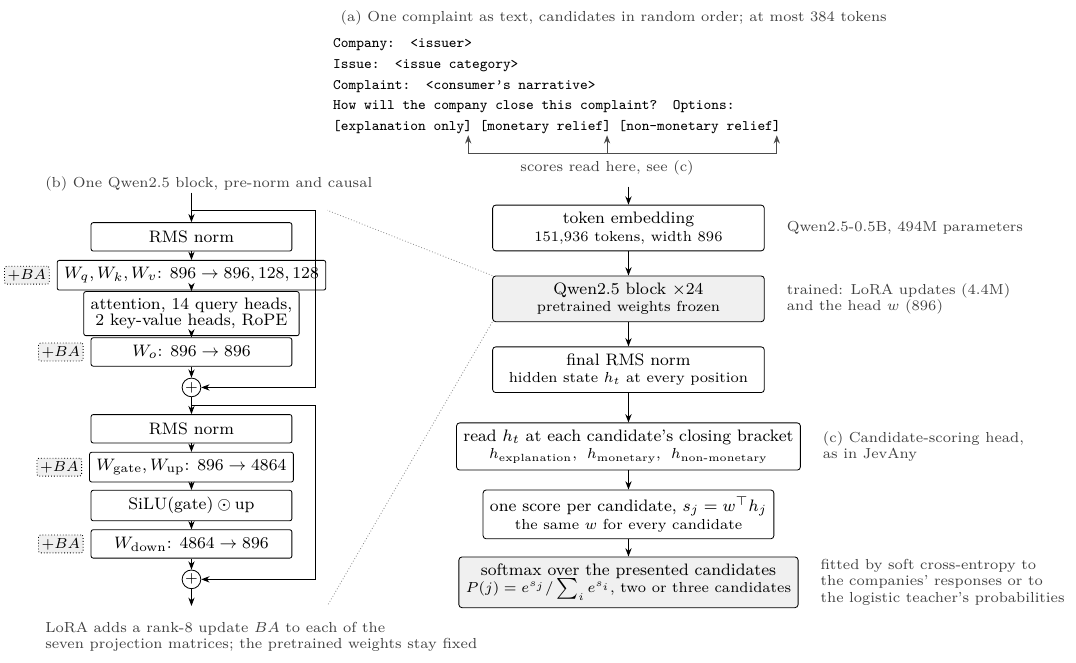}
\caption{The student model for credit card complaints. (a) The input: the company, the issue and the consumer's narrative, followed by the question and two or three candidates in random order. (b) One of the 24 blocks of Qwen2.5-0.5B; the pretrained weights stay fixed, and LoRA trains a rank-8 update $BA$ to each projection matrix. (c) The candidate-scoring head reads the hidden state at each candidate's closing bracket and normalizes the scores over the presented candidates.}\label{fig:cfpbarch}
\end{figure}

The \emph{teacher} is a multinomial logistic regression on word and bigram TF-IDF features of the narrative with indicators for the company and the issue, fitted on the teacher years only. It plays the part of the scorecard a bank already has. Because it was fitted on 2015--2018 and the students on 2019--2023, it is fixed relative to the students' sample, as Section~\ref{sec:limits} requires. Two \emph{zero-shot readings} ask a model to choose one lettered option and renormalize its next-token distribution over the letters: Qwen3-30B-A3B-Instruct \cite{qwen3}, served with vLLM \cite{vllm}, and the method published as a local Jev \cite{garcia}, with Qwen2.5-0.5B-Instruct \cite{qwen25} and a fixed threshold. Each complaint is read with three options \{monetary relief, non-monetary relief, explanation only\} and with two options \{monetary relief, no monetary relief\}, each in two orders, and each format is averaged over its orders. The \emph{students} follow the JevAny design \cite{jevany} (Figure~\ref{fig:cfpbarch}): the input is the company, the issue and the narrative, cut at 384 tokens, followed by the question and the candidates in brackets; Qwen2.5-0.5B \cite{qwen25} with rank-8 LoRA adapters \cite{lora} on its seven projection matrices reads it, and a linear score at each candidate's closing bracket is normalized over the candidates shown. One student is trained on the companies' responses and one on the teacher's probabilities, each on $n=64{,}000$ complaints from 2019--2023 with soft cross-entropy, questions with two or three options in random order and early stopping on 4,000 held-out complaints. Two \emph{fixed representations} sit between the teacher and the students, each with a logistic regression fitted to the companies' responses on the same 64,000 complaints as the students: the mean of the last hidden layer of Qwen2.5-0.5B with its weights frozen, which is the frozen-backbone design of Section~\ref{sec:rep}, and the mean of word2vec vectors \cite{word2vec} trained without labels on the 2015--2018 narratives, so that the representation is fixed before the students' sample is drawn. Recalibration uses the robust logit curve of Tasche \cite[eq.~A.8]{tasche}, $p(x)=1/[1+\exp\{-\alpha-\beta\,\Phi^{-1}(\tilde F_N(x))\}]$, where $\tilde F_N$ is the distribution function of the forecast among complaints without monetary relief, averaged over its left- and right-continuous versions. The parameters are chosen by quasi moment matching \cite[appendix~A.2]{tasche}: the curve reproduces the relief rate of the 5,000 calibration complaints from 2024 and, through the accuracy ratio implied by the curve, their observed accuracy ratio. The curve depends on the forecast only through its rank, so, like the temperature scaling of Corollary~\ref{cor:recal}, it leaves the ranking unchanged. All models are evaluated on the same 10,000 test complaints, of which 16.9\% ended in monetary relief. The code for this section, including the recalibration (\texttt{recal.py}), is in the \texttt{cfpb} directory that accompanies \texttt{experiments.py}; the CFPB files are listed there with their checksums.

\subsection{Measurement without $\eta$}
For any forecast $q$, $\E[(q-Y)^2]=\E[\{q(X)-\eta(X)\}^2]+\E[\eta(X)\{1-\eta(X)\}]$, and the second term does not depend on the model, so the difference in Brier score between two models on the same complaints estimates their difference in excess risk. We report 95\% intervals from a paired bootstrap over complaints. For calibration we use the identity behind Turing's check of weight-of-evidence calculations, $\E(1/F\mid H)=1$ \cite[secs.~7--8]{good}; \cite[Sec.~1.4]{burakov} derives it and notes that a forecast that always gives the base rate satisfies it: with $V=(1-q_Y)/q_Y$ the odds against the realized outcome, a forecast with $\E[Y\mid q]=q$ has $\E[V\mid q]=1$. The decision rule escalates when $q>c$; an unneeded escalation costs $c$ and a missed relief case $1-c$, and we report the cost per 1,000 complaints. Coherence under the refinement of ``no monetary relief'' into its two subevents is measured by the mean absolute difference, in decibans, between the log-odds of monetary relief under the two candidate sets; condition \eqref{eq:coherence} requires zero.

\begin{table}[ht]
\centering\small\setlength{\tabcolsep}{4pt}
\caption{Forecasts of monetary relief on 10,000 test complaints received in 2025--2026; the share with relief is $0.169$. Turing's mean $V$ is 1 for a calibrated forecast. Models trained on the 2019--2023 responses (the two fixed representations and the students) are reported without recalibration; the teacher and Qwen3-30B are recalibrated on 2024. The last column measures the departure from \eqref{eq:coherence}. $^{a}$Uncalibrated forecast; the mean is dominated by confident forecasts that turned out wrong.}\label{tab:cfpb}\vspace{6pt}
\begin{tabular}{lccccccc}
\toprule
& mean & & & Turing & \multicolumn{2}{c}{cost per 1,000} & departure \\
\cmidrule(lr){6-7}
Model & forecast & Brier & AUC & mean $V$ & $c=0.2$ & $c=0.3$ & (decibans) \\
\midrule
Constant at the 2024 rate & $0.171$ & $0.141$ & $0.50$ & $0.99$ & $135.4$ & $118.4$ & \\
Letter reading, 0.5B, as published & $0.262$ & $0.160$ & $0.48$ & $0.95$ & $164.3$ & $165.1$ & $5.0$ \\
Qwen3-30B zero-shot & $0.509$ & $0.364$ & $0.73$ & $219^{a}$ & $100.2$ & $136.7$ & $17.5$ \\
Qwen3-30B, recalibrated & $0.204$ & $0.137$ & $0.73$ & $0.96$ & $100.7$ & $116.0$ & \\
Logistic teacher, recalibrated & $0.183$ & $0.124$ & $0.77$ & $1.12$ & $90.0$ & $100.1$ & \\
word2vec + logistic & $0.175$ & $0.123$ & $0.77$ & $1.05$ & $90.7$ & $103.7$ & \\
Frozen Qwen2.5-0.5B + logistic & $0.186$ & $0.121$ & $0.79$ & $0.95$ & $85.4$ & $99.8$ & \\
Student on the teacher & $0.143$ & $0.121$ & $0.78$ & $1.32$ & $89.6$ & $100.4$ & $0.20$ \\
Student on outcomes & $0.194$ & $0.119$ & $0.80$ & $1.70$ & $82.7$ & $97.3$ & $0.69$ \\
\bottomrule
\end{tabular}
\end{table}

\begin{figure}[ht]
\centering
\includegraphics[width=\linewidth]{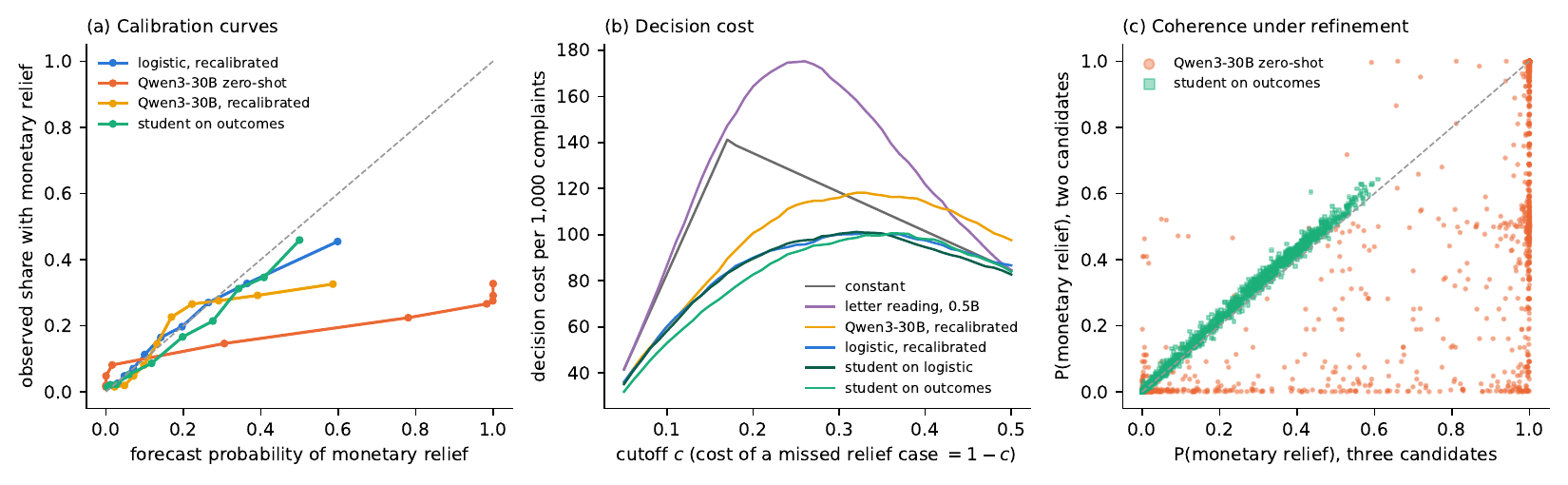}
\caption{Credit card complaints, 10,000 test complaints. (a) Observed share with monetary relief against the forecast, by deciles of each forecast. (b) Decision cost per 1,000 complaints against the cutoff $c$. (c) Probability of monetary relief under the three-option and the two-option candidate sets for 1,500 complaints; a model satisfying \eqref{eq:coherence} lies on the diagonal.}\label{fig:cfpb}
\end{figure}

\subsection{Results}
\emph{Teacher floor.} The student trained on the teacher's probabilities ends where its teacher is (Table~\ref{tab:cfpb}). Its decision cost at $c=0.2$ differs from the teacher's by $-0.4$ per 1,000 (interval $-3.1$ to $2.1$), and its Brier score is lower by $0.0029$ ($0.0015$ to $0.0044$). The student trained on outcomes goes past the teacher: its Brier score is lower by $0.0049$ ($0.0026$ to $0.0073$), its AUC is $0.80$ against $0.77$ and its decision cost at $c=0.2$ is lower by $7.3$ per 1,000 ($3.8$ to $10.9$), or 8\%. This is the pattern of Proposition~\ref{th:teacher} and of \eqref{eq:triangle} at a sample size past the crossover. Against the teacher-trained student, the outcome student's Brier score is lower by $0.0020$ with an interval from $-0.0041$ to $0.0001$, so the separation of the two students at this $n$ rests on one training run each.

\emph{Fixed representations.} The frozen Qwen2.5-0.5B backbone with a logistic head, trained on outcomes, has a decision cost lower than the teacher's by $4.6$ per 1,000 at $c=0.2$ ($1.4$ to $7.9$), and its Brier score is lower by $0.0022$, with an interval from $0.0000$ to $0.0044$ that reaches zero; word2vec does no better than the teacher on either measure. Adapting the same backbone with LoRA lowers the Brier score by a further $0.0027$ ($0.0011$ to $0.0044$) and the decision cost by $2.7$ per 1,000, an interval from $-0.2$ to $5.4$ that just includes zero. The frozen representation therefore discards some signal about the decision that adaptation recovers, the representation loss $B_A$ of Proposition~\ref{th:rep}. The training target matters more than the backbone here: the frozen backbone trained on outcomes has a decision cost lower by $4.2$ per 1,000 ($0.9$ to $7.6$) than the LoRA student trained on the teacher.

\emph{Recalibration.} Read zero-shot, Qwen3-30B ranks complaints usefully (AUC $0.73$) but puts the probability of monetary relief at $0.51$ on average, and in the three deciles it rates highest, where its forecast is close to 1, between 28\% and 33\% of complaints received relief (Figure~\ref{fig:cfpb}a). Its calibration slope \cite{cox}, the coefficient of a logistic regression of the outcome on the logit of the forecast in 2024, is $0.128$, so its log-odds are about eight times too extreme; the slope describes the miscalibration, and the recalibration itself uses the rank-based curve above. Recalibration lowers its Brier score from $0.364$ to $0.137$ and leaves its AUC unchanged, and the recalibrated model remains worse than the teacher by $0.0133$ in Brier score ($0.0101$ to $0.0166$) and by $10.7$ per 1,000 in decision cost at $c=0.2$ ($6.6$ to $15.1$). As Corollary~\ref{cor:recal} states for the probit model, recalibration corrects the scale of a forecast and leaves the excess that comes from its ranking.

\emph{Calibration diagnostics.} Turing's mean $V$ is close to 1 for the recalibrated Qwen3-30B ($0.96$), for the published letter reading, which ranks complaints no better than chance ($0.95$), and for the constant forecast ($0.99$). The best model, the student trained on outcomes, gives $1.70$, because the 1\% of complaints with the largest $V$, confident forecasts that turned out wrong, make up two thirds of its sum. The check measures calibration only, and a validator who relied on it would accept the constant forecast and reject the best model. This is the statement after Corollary~\ref{cor:recal} that calibration diagnostics do not detect the residual \eqref{eq:residual}.

\emph{Coherence.} Qwen3-30B, which reads all candidates in one context, departs from \eqref{eq:coherence} by $17.5$ decibans on average: its probability of monetary relief averages $0.51$ with three candidates and $0.31$ with two, and many complaints receive a probability near 1 under one set and near 0 under the other (Figure~\ref{fig:cfpb}c). The published letter reading departs by $5.0$ decibans, and reversing the order of its options moves its answer by $0.40$ on average. The trained students depart by $0.20$ and $0.69$ decibans, and reversing the order moves their answers by $0.007$ and $0.012$. The students were trained with candidate sets and orders drawn at random, and they come close to \eqref{eq:coherence} although their architecture does not impose \eqref{eq:logsum}.

\section{Scope and limitations}\label{sec:limits}
A deterministic linear teacher corresponds to $\tau=0$ and is covered. A deterministic nonlinear teacher has residuals that are uncorrelated with the features in the population but in general depend on $X$ and are heteroscedastic given $Z$, so the independence step in Theorem~\ref{th:unified} fails. A teacher trained on records that overlap the student's sample makes $\delta$ and the teacher errors depend on that sample. Training LoRA adapters, as in the JevAny models, makes the representation depend on the labels and is not covered.

Fitting an intercept changes the degrees of freedom. With centered Gaussian predictors and an estimated intercept, the centered Gram matrix is distributed as $W_d(\Sigma,n-1)$, so the slope-risk factors $d/(n-d-1)$ and $r/(n-r-1)$ become $d/(n-d-2)$ and $r/(n-r-2)$, for $n>d+2$ and $n>r+2$ respectively; the intercept estimate contributes a further term to the prediction error.

The probit results describe fixed teacher forecasts and population recalibration. They do not give the finite-sample distribution of a fitted nonlinear classifier or of a temperature estimated on a finite calibration set.

Section~\ref{sec:cfpb} carries out the second of the following tests on credit card complaints, a version of the first, and, with the frozen-backbone and word2vec representations, a version of the third. The first perturbs the serialization of a fixed tabular record (field order, number formatting and label wording) and measures the variation of the output probability within an observation. The second divides one candidate into mutually exclusive subcandidates and measures the departure from \eqref{eq:coherence}. The third compares a candidate-scoring model with a frozen pretrained embedding followed by a conventional calibrated classifier or kernel method, both trained on the same observations and evaluated with proper scores on held-out outcomes over paired splits and seeds. The present analysis gives the statistical tradeoffs that such a comparison involves but does not predict its outcome. In Section~\ref{sec:cfpb} the perturbation is the wording and order of the candidates, and the comparison is between a logistic teacher, zero-shot readings, two fixed representations with logistic heads and LoRA students on the same complaints. That application uses one training run per student and one training size, so the crossover sample size was not located; its label is a company's decision, which encodes the company's policy as well as the content of the complaint; and its models lie outside the Gaussian, probit and fixed-representation assumptions, so it tests the direction of the results; their values do not carry over.

Finally, predicting an outcome and selecting an action are different problems. A recommendation label may encode a human policy, a teacher's judgment or an optimization rule, and the appropriate target and loss depend on which of these is to be predicted. When actions change outcomes, causal assumptions are required in addition to the statistical ones used here.

\section{Conclusion}\label{sec:conclusion}
Under Gaussian surrogates the three training ingredients documented for candidate-scoring models have exact consequences. Training on teacher targets leaves a risk floor equal to the teacher's coefficient bias, so it improves on outcome supervision only below a crossover sample size. A fixed representation trades the representation loss $B_A$ against the estimation variance; it also projects the teacher bias onto the retained directions and moves the unrepresented teacher signal into the variance. The optimal teacher weight depends on the sample size, the teacher distortion and the correlation of the target errors. In the probit model, temperature scaling corrects a teacher whose coefficient has the right direction, and a residual excess, given in closed form, remains when the direction is wrong. The regression formulas assume a Gaussian design; under a multivariate-$t$ design the simulated risk exceeds them by up to $35\%$ at $n=12$, and the discrepancy shrinks as $n$ grows. On credit card complaints the direction of each result held outside these assumptions: a student trained on a teacher stayed at the teacher's decision cost while a student trained on outcomes went past it, recalibration left a large zero-shot model worse than a logistic regression, Turing's calibration check accepted a constant forecast and rejected the best model, and a model reading all candidates in one context departed from the coherence condition by 17.5 decibans.

For practitioners the results suggest five checks.
\begin{enumerate}
\item \emph{Training target.} A model trained on teacher or reference probabilities estimates the teacher's conditional distribution. Its error against outcomes includes a teacher bias that more student data does not remove (Proposition~\ref{th:teacher}) and that agreement with the teacher does not reveal \eqref{eq:triangle}.
\item \emph{Evaluation.} Proper scores should be computed on realized outcomes. The model should be compared with a conventional classifier fitted to the same outcomes and with one fitted to the same teacher probabilities, so that the effect of the architecture is separated from the effect of the training target.
\item \emph{Pretrained representations.} A fixed representation reduces estimation variance. It lowers the risk only while the signal it discards is below the threshold \eqref{eq:crossover}, and the threshold decreases as the sample grows.
\item \emph{Calibration.} Recalibration corrects the scale of a forecast. A calibrated forecast can retain the excess risk \eqref{eq:residual} of a wrong coefficient direction, and calibration diagnostics do not detect it.
\item \emph{Coherence.} The same event should receive the same probability under different candidate sets \eqref{eq:coherence}, which can be tested by posing the event under several sets.
\end{enumerate}

A model that predicts which candidate an annotator, a teacher or a past policy would select has not been shown to select the action with the best outcome. Establishing that requires a model of outcomes given actions and, when actions change outcomes, causal assumptions.

\end{document}